\documentclass[11pt,a4paper]{article}
\usepackage[hyperref]{emnlp2018}
\usepackage{times}
\usepackage{latexsym}
\usepackage{graphicx}
\usepackage{url}

\aclfinalcopy 

\title{Neural Multi-Task Learning \\ 
for Citation Function and Provenance}

\author{Xuan Su \qquad
  Animesh Prasad \qquad
  Kazunari Sugiyama \qquad
  Min-Yen Kan \\
  School of Computing, National University of Singapore \\
  13 Computing Drive, Singapore\\
  \tt \{suxuan,animesh,kazunari,kanmy\}@comp.nus.edu.sg
  }

\date{}

\begin{document}
\maketitle
\begin{abstract}
Citation function and provenance are two cornerstone tasks in citation analysis. Given a citation, the former task determines its rhetorical role, while the latter locates the text in the cited paper that contains the relevant cited information. We hypothesize that these two tasks are synergistically related, and build a model that validates this claim. For both tasks, we show that a single-layer convolutional neural network (CNN) outperforms existing state-of-the-art baselines. More importantly, we show that the two tasks are indeed synergistic: by jointly training both of the tasks in a multi-task learning setup, we demonstrate additional performance gains. Altogether, our models improve the current state-of-the-arts up to 2\%, with statistical significance for both citation function and provenance prediction tasks.
\end{abstract}

\section{Introduction} \label{sec:Intro}
In academia, citations are an important tool that helps acknowledge the intellectual credit to prior sources of knowledge. In the domain of computer science alone, the {\it Times Higher Education}~\cite{citationaverages} stated that papers published post-2000 received 
7.17 citations on average. 
This prevalence of citations leads bibliometric researchers to examine citations and their accompanying text as a rich source for understanding how it facilitates  networks and communications in scientific discourse.

Why do authors make citations?  What kind of rhetorical role do they play? Reflecting on our first citing sentence in this paper above, it is clear that the citation is a reference to general statistics from a trustworthy, expert source.  With this citation, we establish the reliability of the information used. This leads us to the task of citation function classification, where a system assigns one out of a set of predefined rhetorical roles to a given citation. 

To be specific, the task of citation function aims at determining the function that a given citation plays in its context. For instance, in machine learning literature, it is common to see comparison of performance of different methods. In such cases, citations to the performance scores are considered to serve a {\it compare} and {\it contrast} purpose. Yet another example is citations which refer to weaknesses of the cited paper. This may occur in cases where the citing paper is an attempt to improve on top of the cited approach. These are just two cases on how citations may serve various roles in their contexts; in Section~\ref{sec:methodology}, we define rigorously the four citation functions considered in this paper.

Additionally, readers are not bound by the frame of the citing paper; they may trace through the citation to the cited paper -- the original paper containing the purported cited information.   This introduces the second task of interest: citation provenance identification, the task of identifying the cited information in the cited paper, corresponding to a given citation.

To be specific, the task of citation provenance aims to recognize texts in the cited paper that reflect the content of a given citation. For example, citations to performance of models may refer to the numeric, tabulated scores in the target paper, or simply to a sentence containing the detailed method evaluation. Citations to weaknesses 
may be reflected in sentences in the cited paper that unveil future directions or improvements on the existing work. In our paper, we model this task as a binary classification task. In other words, given a citation, for each given text fragment in the cited paper, we label it as 
either {\it provenance} or {\it non-provenance}.

Most existing approaches to the above two tasks employ rule-based or conventional machine learning models. These methods have a clear disadvantage: a large amount of manual effort is required to obtain rules or features. This implies that such models may only be applied to restricted target domains where they were developed for, since the rules or features are bespoke.

Recent advances in deep learning allow the cumbersome domain-specific engineering processes to be bypassed entirely.  Words are transformed into embeddings: high-dimensional numeric vectors that encode a unit of language's semantics through an analysis of its distributional contexts. 
Using sufficient degrees of freedom in modeling through its parameters and relevant training methodologies, current neural network models can represent a large class of implied features without laborious manual engineering. 
Our work re-examines these two key tasks of scholarly document processing under the auspice of deep learning.

Furthermore, although researchers have approached these two tasks separately using neural networks, no attempt has been made to examine the pair of tasks in tandem. Intuitively, there is a correlation between the function and provenance labels that we would assign to a given citation. Knowing the function of a given citation may then help determine whether a text fragment is its corresponding provenance, and vice versa. For instance, it is unlikely that a citation expressing approval of another paper refers to a section in the target paper that in fact reveals its own drawback. Similarly, if a citation 
is found to be comparing the effectiveness of the cited approach with others, it 
likely references the `Evaluation' 
section, since that is the section where evaluation figures are usually disclosed. This implicit relationship between citation function and provenance prompts us to attempt multi-task learning (MTL)~\cite{caruana1998multitask} for both tasks. Such a learning paradigm is said to be able to exploit commonalities and differences across tasks, which usually leads to improved results. 


Our contributions are summarized as follows\footnote{Code and data available at \url{https://github.com/animeshprasad/citation_analysis.git}}:

\begin{enumerate}
\item We apply a convolutional neural network (CNN)  
to classification of both citation function and provenance. 
To the best of our knowledge, no unified neural network model has been 
applied to both tasks.  We show that a one-layer CNN model surpasses the performance of rich-feature based baselines. 

\item We demonstrate that citation function and provenance are closely 
  related. We hypothesize that multi-task learning exploits the relationship between the two tasks to further enhance the base performance of neural models.  Experimental results verify our hypothesis (\textit{cf.} Section~\ref{sec:evaluation}).

\item As the numbers of instances in each class are skewed in both tasks, we develop a selective crowdsourcing methodology to construct a dataset better suited for supervised learning. We then illustrate our use of batch-wise selective parameter tuning to train the models.
\end{enumerate}


This paper is organized as follows. 
In Section \ref{sec:related_work}, we briefly survey previous studies on citation function and provenance. 
We also review deep learning-based architectures in natural language processing and the multi-task learning framework. In Section~\ref{sec:methodology}, we describe the main methodological approaches, including details of the datasets, formulation of tasks, and details of the models. In Section~\ref{sec:evaluation} and~\ref{sec:discussion}, we present our experimental results and discuss a few classification output examples. Finally, we conclude our work with a summary and future directions 
in Section~\ref{sec:conclusion}.

\section{Related Work \label{sec:related_work}}
In this section, we review related work on both citation function and provenance tasks. We also
briefly review research on neural multi-task learning. 

\subsection{Citation Function}
Research on citation functions dates back to the 1970s. 
Moravcsik and Murugesan~\cite{moravcsik1975some} made the first in-depth study of citation functions. They proposed a series of four questions to classify citations into the following four categories: (1) conceptual versus operational, (2) organic versus perfunctory, (3) evolutionary versus juxtapositional, and (4) confirmative versus negational.  
At the turn of the millennium, 
Garzone and Mercer~\cite{garzone2000towards} constructed 
the first automatic classifier for citation functions. 
This research employed a rule-based grammar exploiting citation's cue words and section information to classify citations into 35 classes.

From then on, many contributions focused on articles from specific domains, and usually utilized manually constructed features. For instance, 
Teufel \textit{et al.}~\cite{teufel2006automatic} focused on computer science domains. They employed a variant of the $k$-nearest neighbor algorithm with rich features based on linguistic cues to classify citations into twelve categories.  
As another example, 
Agarwal \textit{et al.}~\cite{agarwal2010automatically} addressed citations in the biomedical domains. 
They classified citations into eight functions and constructed a Support Vector Machine classifier using a combination of unigram and bigram features. 
Abu-Jbara \textit{et al.}~\cite{abu2013purpose} adopted and re-categorized the 12 classes defined by Teufel \textit{et al.}'s work~\cite{teufel2006automatic} into six classes. The new classification scheme along with new rich surface and linguistic features resulted in marginally better results. 
They also observed that citation functions and sentiments of citation contexts are closely related, and built a classifier for detecting polarity (\textit{i.e.}, author sentiment) in citations. Jha \textit{et al.}~\cite{jha2017nlp} did similar feature-driven analysis with a small dataset created from ten documents, with six classes. 
A common trait among these works is that they explore expert-designed features to classify the functions. Linguistic cues designed to suit a particular taxonomy are commonly used. Though effective, such techniques are highly dependent on the domain and taxonomy \cite{jha2017nlp}.

Recently, the emergence of deep learning has led researchers to attempt neural models for citation analysis. For example, Munkhdalai \textit{et al.}~\cite{munkhdalai2016citation} used a compositional attention network to classify citations, yielding consistently good performance. They classify a highly skewed PubMed citation dataset into sentiment and function groups using a Long-Short Term Memory network. However, we note that these models are designed specifically for the citation function task, without consideration of related, synergistic tasks which may bring further enhancement. We address this research gap in this paper.

\subsection{Citation Provenance}
While researchers have worked on citation function, significantly fewer have addressed citation provenance.  Wan \textit{et al.}~\cite{wan2009designing} analyzed researchers' literature browsing habits and revealed that 
while encountering citations, readers would find it useful if there is a tool to identify important sentences in the cited paper that justify the citation. This hints at the usefulness of an intelligent reading tool with citation provenance support. Low~\cite{wee2011citation} introduced the first automatic tool to identify citation provenance, where the following two-tier approach is employed.
The first tier classifies citations into either {\it general} or {\it specific}, and the second tier identifies the relevant provenance texts for citations marked as {\it specific}. 
This approach is able to determine the cited fragment(s) in the cited paper, given knowledge that a citation is {\it specific}. The task started attracting more attention recently as a pre-processing step for generating faceted comprehensive summaries of scientific documents. In the CL-SciSumm Shared Task 2016~\cite{jaidka2018insights}, 
most systems used traditional features similar to Low's approach~\cite{wee2011citation}. These systems show reasonable performance with wide variance. 
None of the reported systems employed an end-to-end deep learning model for provenance identification, with the closest inspiration to our work being the hybrid system proposed by~\cite{prasad:2017:CL-SciSumm}.

\begin{table*}
\centering
\caption{\label{citation-function-scheme-table} Citation function classification scheme with examples.}
\begin{tabular}{|p{3cm}|p{5cm}|p{5cm}|}
\hline
\textbf{Category} & \textbf{Explanation} & \textbf{Citing Sentence Example} \\ \hline
\hline
Weakness (\textit{Weak})      & The citation points to weaknesses or problems of the cited paper. & Smith's system \textbf{(2010)} fails to take into consideration many other factors. \\ \hline
Compare and Contrast (\textit{CoCo}) & The citation compares or contrasts the results or methodology from the cited paper with another work. & Our results are significantly better than those reported in Joe's work \textbf{(2011)}.  \\ \hline
Positive (\textit{Pos})    & The citation expresses approval of the cited paper. For example, the citing paper adopts an idea, method or dataset from the cited paper, or it shows compliment of the cited paper. & Our system, called BusTUC is \textbf{built upon} the classical system CHAT-80 \textbf{(Warren and Pereira, 1982)}. \\ \hline
Neutral (\textit{Neut})    & The citation serves a neutral purpose: background, mere mentioning, etc; or its function is not decidable. & At the University of Trondheim (NTNU), two students made a Norwegian version of CHAT-80, called PRAT-89 \textbf{(Teigen and Vetland, 1988)}. \\ \hline
\end{tabular}
\end{table*}

\begin{table*}
\centering
\caption{\label{citation-prov-scheme-table} Citation provenance with examples.}
\begin{tabular}{|c|p{6cm}|p{7cm}|}
\hline
\textbf{Category} & \textbf{Citing Sentence Example} & \textbf{Target Fragment} \\ \hline
\hline
\textit{Prov}                 & Consequently, current anaphora resolution methods rely mainly on constraint and preference heuristics, which employ morpho-syntactic information or shallow semantic analysis (see, for example, \textbf{Mitkov [1998]}). & It makes use of only a part-of-speech tagger, plus simple noun phrase rules (sentence constituents are identified at the level of noun phrase at most) and operates on the basis of antecedent-tracking preferences (referred to hereafter as ``antecedent indicators''). \\ \hline
\textit{Non-Prov}                 & Consequently, current anaphora resolution methods rely mainly on constraint and preference heuristics, which employ morpho-syntactic information or shallow semantic analysis (see, for example, \textbf{Mitkov [1998]}). & Given that our approach is robust and returns an tecedent for each pronoun, in order to make the comparison as fair as possible, we used CogNIAC's ``resolve all'' version by simulating it manually on the same training data used in evaluation B above.  \\ \hline
\end{tabular}
\end{table*}

\subsection{Neural Multi-Task Learning}
Multi-task learning is a machine learning technique that has been useful in many real-world applications. Caruana~\cite{caruana1998multitask} has demonstrated that multi-task learning is able to exploit the information contained in the training signals of related tasks, to improve learning and generalization for a given task. Combined with artificial neural nets, multi-task learning has been successfully applied to a number of natural language processing tasks, such as 
part-of-speech tagging~\cite{Plank16}, comparison of task relationships~\cite{Bingel17}, sequence labeling tasks~\cite{rei:2017:Long}, and so on.

\section{Proposed Method \label{sec:methodology}}

We first formally define our terminology and the tasks of citation function and provenance classification. To the best of our knowledge, there is no decently-sized, publicly available dataset to learn citation function.  Thus, we next describe how we construct our own citation function dataset. We also provide descriptions of the citation provenance dataset taken from CL-SciScumm 2016. Then, we describe our unified architecture based on a convolutional neural network, and explain how we apply multi-task learning for joint training of both tasks.

\subsection{Definitions}
A \textit{citation} is a formal attribution to prior work,
usually explicitly marked by a conventional {\it citation marker},
which can contain author name(s) and/or the year of publication.  
We deem the {\it citing sentence} as the single sentence that
physically contains the citation marker. 
It provides important
information about how the citation is used. Related to this, 
{\it citation context} --- the logically entailing context of the
citation --- gives readers more insights into the article flow
surrounding the citation.  
Generally, the judgment of the citation context may be subjective 
and is a variable-length span 
which can be as short as a noun phrase, or as long as a paragraph.  
In our work, we define it as a set of three sentences, including citation sentence and its previous and following sentences. 

\subsection{Classification Schemes}
{\bf Citation Function} \\ 
We adopt the classification scheme in~\cite{yulianto2012citation}, 
which in turn is based on a simplification of the seminal classification scheme 
in~\cite{teufel2006automatic}. 
This scheme consists of the following four classes: ({\it Weak})ness, Compare and Contrast ({\it CoCo}), ({\it Pos})itive, and ({\it Neut})ral.  The scheme is 
general 
({\it i.e.}, non domain-specific) and  
indicative of the sentiment of the
citation~~\cite{abu2013purpose, jha2017nlp}. 
Table~\ref{citation-function-scheme-table} summarizes this scheme. \\

\noindent
{\bf Citation Provenance} \\
We use a binary classification scheme, 
{\it Prov} and {\it Non-Prov}. 
Given a citation context and a target fragment, the fragment is classified as {\it Prov} if it contains evidence for the cited information. Otherwise, it is labelled as {\it Non-Prov}. Table~\ref{citation-prov-scheme-table} 
summarizes this scheme. 

\subsection{Dataset \label{sec:dataset}}
\noindent
{\bf Citation Function} \\
Although there has been significant prior work, 
a sizeable dataset for learning citation functions 
has not been made publicly available. Therefore, we manually annotate data for citation function to construct our own corpus
(see Table~\ref{citation-function-dataset-table}). Citation contexts are
taken from randomly-sampled articles in the ACL Anthology Reference
Corpus~\cite{bird2008acl}. 
We used the CrowdFlower\footnote{\url{https://www.crowdflower.com/}}
platform to crowdsource annotations under the scheme shown in
Table~\ref{citation-function-scheme-table}, where annotation quality is 
controlled by the platform. 


We collect data in two rounds in order to enrich the quality of the dataset. The first data collection round includes 1040 citation contexts taken from the ACL Corpus. Annotation results reveal that the distribution of citation functions is highly skewed, with $79.40\%$ of instances labeled {\it Neut}. 
To alleviate the extreme ratio between \textit{Neut} and other classes, we next selectively crowdsource additional instances 
for the three minority classes. A more balanced class ratio not only helps us make better supervised models, but also facilitates benchmarking evaluation. For many downstream applications, it is often more desirable to have higher prediction performance on the minority, non-\textit{Neut} classes. We employ 
the following two strategies for selective crowdsourcing.
\begin{enumerate}
    \item We analyze
the sentiments of the annotated citing sentences using Google Cloud
Language
API\footnote{\url{https://cloud.google.com/natural-language/}} to
establish representative statistics. In Google's framework, sentiment
scores range from -1 (very negative) to 1 (very positive).  Scores
above 0 exhibit positive emotions and larger absolute values imply
stronger valence. 
We observe that the average sentiment scores for \textit{Weak} 
and \textit{Pos} instances are -0.367 and 0.241, respectively. 
The remaining \textit{CoCo} and \textit{Neut}
classes have mean sentiment values close to 0:
-0.047 and 0.105, respectively. 
Since the sentiments of most instances from
each class are expected to cluster around their class mean, we set 
sentiment scores to $\pm$0.6 as cut-offs, and include instances with
such strong valence from unused articles as candidates for annotation, 
to produce a larger proportion of {\it Weak} or {\it Pos} annotations.

\item Manual examination of the annotated
instances shows that particular linguistic cues appear more often in
certain classes. For example, the word ``contrast'' frequently appears in {\it
  CoCo} citations, while more {\it Pos} instances include the phrase
``we use''.  The complete list of cues is: ``Contrast'', ``Comparable'' and ``Similar to'' for {\it CoCo}; ``We use'', ``We have used'', ``We adopt'', ``We have adopted'', ``I use'', ``I have used'', ``I adopt'', ``I have adopted'', ``We follow'', ``I follow'' for {\it Pos}. We include instances with these selected linguistic
cues into our dataset for annotation.
\end{enumerate}

Instances collected using these two strategies are then similarly 
hosted on CrowdFlower for the second round of annotation. In total, $392$
instances are annotated, 
resulting in $207$ minority instances. 
The percentage of minority instances is $52.81\%$, much more than
the $20.60\%$ proportion of the first round.
This shows that our selective data collection approach is effective. \\

\begin{table}
\centering
\caption{Statistics on citation function dataset.}
\label{citation-function-dataset-table} 
\begin{tabular}{|c|c|c|c||c|}
\cline{1-5}
\bf Weak & \bf CoCo & \bf Pos & \bf Neut & \bf Total \\ 
\cline{1-5}
\cline{1-5}
31 & 95 & 295 & 1011 & 1432 \\
\cline{1-5}
\end{tabular}
\end{table}

\begin{table}
\centering
\caption{Statistics for our citation provenance dataset.}
\label{citation-provenance-dataset-table}
\begin{tabular}{|c|c||c|}
\cline{1-3}
\bf Prov & \bf Non-Prov & \bf Total \\ 
\cline{1-3} 
\cline{1-3}
608 & 885 & 1493\\
\cline{1-3}
\end{tabular}
\end{table}

\noindent
{\bf Citation Provenance} \\
In the CL-SciSumm Shared Task 2016~\cite{jaidka2018insights}, Task~1A requires participants to develop systems to identify the spans of text in the cited paper that most accurately reflect the citation, for each given citing context.  They developed a dataset  
specific to system evaluation for this sub-task. 
We directly use their public dataset, 
as it is exactly relevant to our purpose. 
However, due to the nature of the task, the dataset contains positive ({\it Prov}) instances only. Thus, we must supplement and source for our own {\it Non-Prov} instances.  In our manual examination of cited papers, we observe that the vast majority of sentences are actually 
{\it Non-Prov} instances. In other words, sentences in cited papers that at least partially reflect the content of a given citation are a small minority.
As we aim at keeping class balance to construct better model 
and to produce a classifier
that can distinguish informative fragments from trivial ones, we do not utilize all unannotated sentences as {\it Non-Prov}.
Rather, we sample such negative instances by randomly taking three instances from each paper, assuming that unassessed instances constitute members of {\it Non-Prov}. We use three negative instances to achieve approximate class balance (see Table~\ref{citation-provenance-dataset-table}). Table~\ref{citation-prov-scheme-table} shows an example of citation provenance.

\subsection{Model}

While the previous state-of-the-arts often use feature engineering-based systems (detailed later in Section~\ref{sec:baselines}), we use a simple convolutional neural network (CNN) layer as our basic learning
model. We now explain how we adapt the CNN to model text classification for both citation function and provenance tasks.\\

\begin{figure*}[t]
\centering
\includegraphics[width=0.8\linewidth]{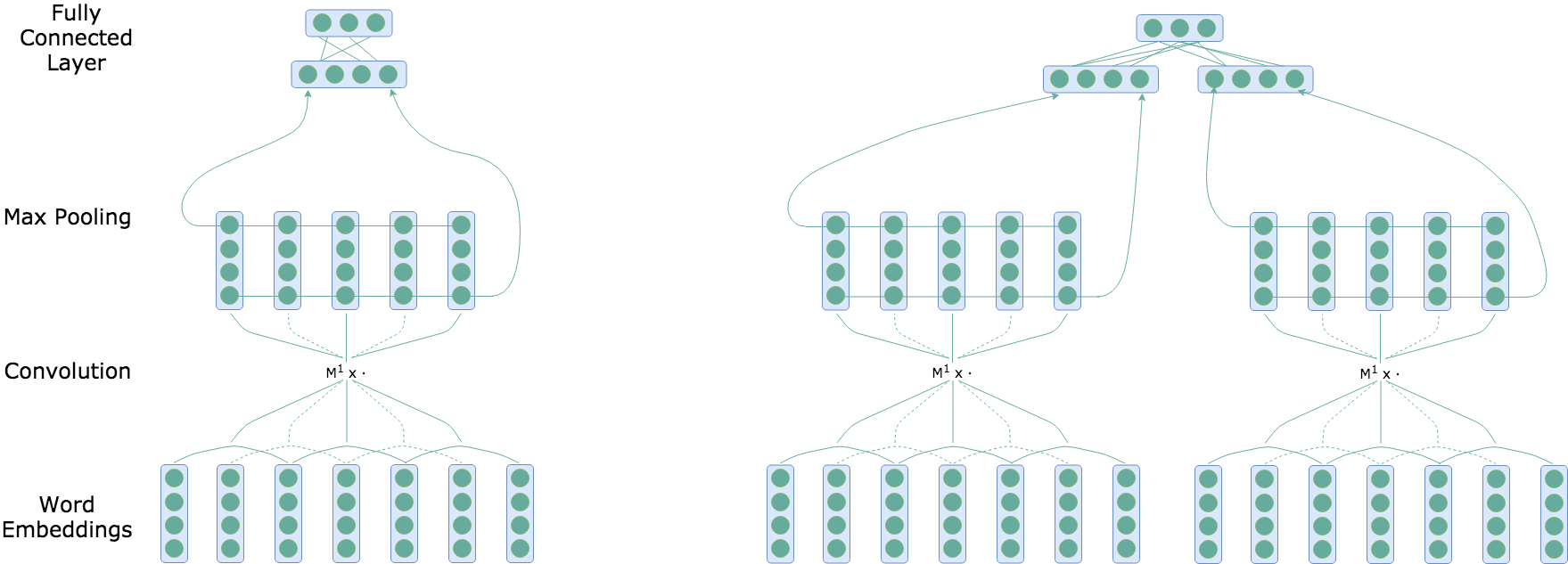}
\caption{CNN (left) and dCNN (right) architectures for citation function for citation provenance respectively.} \label{fig:cnn2cnn}
\end{figure*}

\begin{figure*}[t]
\centering
\includegraphics[width=0.6\linewidth]{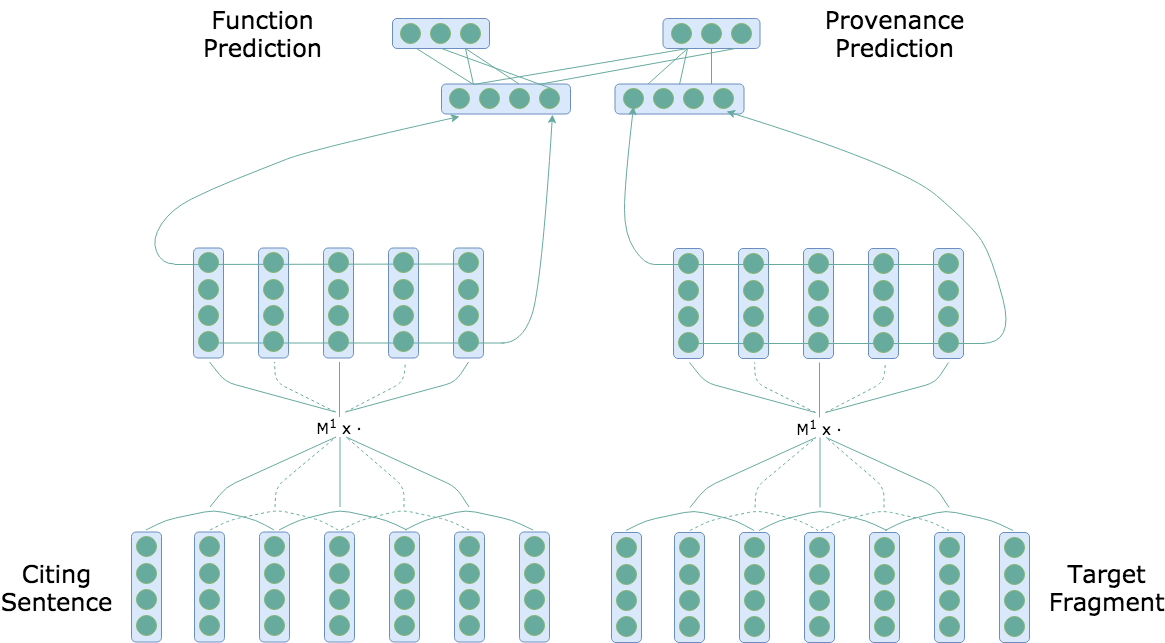}
\caption{Our proposed MTL architecture.} \label{fig:mtl}
\end{figure*}

\noindent
\textbf{Convolutional Neural Network (CNN)}\\
Figure~\ref{fig:cnn2cnn} (left) shows the  CNN architecture. This network consists of the following four layers:
\begin{itemize}
\item{Word Embeddings Layer} \\
This layer performs basic pre-processing. Specifically, the input text sequence is padded for length differences, converted to word indexes, and mapped to word embeddings. We use GloVe vectors, pre-trained from a 2014 dump of Wikipedia\footnote{\url{https://nlp.stanford.edu/data/glove.6B.zip}}, with vector dimension of 100. The output is a 2D matrix of size $dim \times w$, 
where $dim$ and $w$ are the embedding dimension (100) and the number of words in the sequence, respectively. 

 \item{Convolution Layer}  \\
This layer applies a series of convolution-like operations on successive word windows of a fixed size ($n=5$). Each operation performs a cross-correlation, multiplying a learned weight matrix with a word window, yielding a single column in the resulting matrix. 

\item{Max Pooling Layer}  \\
This layer squashes the two-dimensional matrix into a one-dimensional vector. The row-wise max pooling discards word (embeddings) that do not contain signal pertinent to the classification decision. 

\item{Fully Connected Layer} \\
This final layer consists of non-linear transforms that combine the component evidence to yield probablistically interpretable class predictions: in our case, the function or provenance labels.  As is standard in neural supervised classification, we employ a cross-entropy loss and softmax to generate class predictions. The softmax function is of the form \\ $s(x_i) = \frac{e^{x_i}}{\sum_{i} e^{x_i}}$, which converts the values in the given vector into probabilities.
\end{itemize}

\noindent
While many existing works claim that adding hand-crafted features to neural nets further boosts classification performance, we do not augment our convolutional model with such features as we consider them to be domain-specific. \\

\noindent
\textbf{Double CNN} \\
The  CNN performs classification on a single text sequence.  For citation provenance, both the citation context and the cited paper fragment need to be jointly considered to make a classification decision.  While concatenation of both sources is possible and simple, the two sources represent logically separate evidence, where the cited fragment should entail the citation context, not vice versa. This motivates us to re-design the CNN architecture. 
We use the double CNN (hereafter, ``dCNN'') architecture for this task (similar to \cite{bromley1994signature}), where two CNNs accept and process the two inputs separately, but combine at the fully-connected layer to generate class predictions (see Figure~\ref{fig:cnn2cnn}, right). We make use of such a dual network as it has been previously verified to work well for tasks that require computing over two separate text sequences ({\it e.g.} \cite{yin2015convolutional}).\\

\noindent
\textbf{Multi-Task Learning (MTL)}\\
A central claim of our work is that citation function and provenance are two interrelated tasks: knowing the function of a given citation may help determine whether a text fragment is its corresponding provenance, and vice versa. The relationship between citation function and provenance motivates us to apply multi-task learning to both tasks. In general, MTL has been shown to improve learning efficiency and prediction accuracy for tasks involved in the training process~\cite{caruana1998multitask}. 

Our work requires input of the citing sentence in both tasks. 
Therefore, it is natural to share parameters that work on the citing sentence. In other words, the dCNN model is employed, with the left half dealing with both citation function data and the citing sentences from citation provenance data. The network working on the text fragment is separate and does not share parameters with the other part. Figure \ref{fig:mtl} illustrates our MTL architecture. We 
evaluate this shared model against the CNN-based models 
on our dataset.

MTL trains all tasks of interest jointly, in order to improve generalization errors. Generally, MTL architectures share some part of the network and introduce task-specific losses over the same set of features. Multiple losses can learn more generalized representation of the features. Works adopting MTL mentioned in Section \ref{sec:related_work} all have their lower layers of the network completely shared, where strong evidence from texts is used by all the tasks. No existing works have employed multi-task learning models to incorporate information from weakly related evidences. In our model, the function half of the multi-task model is trained without aligned provenance data. That is, instances in our dataset are annotated with only either function or provenance labels.
Our multi-task learning is still able to effectively incorporate weak evidence from such unaligned data samples. We note that not all multi-task setups are effective: finding a proper group of tasks that benefit from the multi-task setup implies a deep relation between the tasks~\cite{Alonso17}. 

\section{Experiments\label{sec:evaluation}}
We first introduce the supervised baselines implemented for comparative 
evaluations. Then, we briefly describe our implementation of the neural models. Lastly, we report our experimental results.

\subsection{Supervised Baselines\label{sec:baselines}}
\noindent{\bf Citation Function} \\
We reimplement the classifier previously proposed by 
Yulianto~\cite{yulianto2012citation}. The original classifier achieved $F_1$ score of around $68\%$ when applied to the dataset in his paper. It employs two types of features: local and global. 
\textit{Local} features are directly derived from the citation context, while \textit{global} features are derived from the citing paper.

\begin{enumerate}
\item \textit{Unigram Feature} \\ 
This feature is derived from the citation context using the unigram language model. It measures how often each word appears in the contexts. We apply dimensionality reduction to reduce the size of the feature to 300.

\item \textit{Citation Density} \\ 
This feature is defined as the number of citation markers appearing in the citation context.

\item \textit{Year Difference} \\ 
This feature measures the difference between the publication years of the citing and cited papers.

\item \textit{Citing Location} \\ 
This feature is the location of the citing sentence within the whole citing paper. Specifically, we define citing location as the proportion of words appearing before the citing sentence over the whole article. Thus, this value is a decimal number in the range $[0, 1)$.

\item \textit{Citation Frequency of Cited Paper} \\ 
This feature indicates the number of times the cited paper is referenced in the citing paper.

\item \textit{Number of Other Citations with the Same Author} \\ 
This feature denotes the number of other citations that share the same author as the cited paper.

\item \textit{Self Reference} \\ 
This is a binary feature representing whether there is any common author in the citing and cited papers.\\
\end{enumerate}

Given the dataset, for each citation, we feed it into our feature extractor to obtain the above seven features. We then apply dimensionality reduction to reduce the size of the unigram vector down to 300. Next, we combine this vector with other features, and finally apply a multinomial Na\"ive Bayes classifier to obtain the predicted labels. These features are commonly employed in most of the prior work discussed in Section~\ref{sec:related_work}.\\

\noindent{\bf Citation Provenance} \\
We adapt the classifier previously implemented in \cite{wee2011citation}. Given a citation context and a target fragment, 
We construct a decision tree model by extracting the following features. 

\begin{enumerate}
\item \textit{Surface Matching} \\ 
This feature measures the number of common words appearing in both the citation context and the fragment.

\item \textit{Number Matching} \\ 
This feature counts the common decimal numbers appearing in both the citation context and the fragment.

\item \textit{Bigram Matching} \\ 
This feature measures the number of common bigrams appearing in both the citation context and the fragment.

\item \textit{Cosine Similarity} \\ 
This feature is a common measurement of document similarity with words as the basic unit. In our work, it represents the similarity between the citation context and the fragment.
\end{enumerate}

This model is mainly based on the idea that the citation context and its corresponding provenance are likely similar to each other in terms of surface word comparisons. These features broadly cover the features used by most models in CL-SciSumm 2016~\cite{jaidka2018insights}.

\begin{table*}
\centering
\caption{Performance of citation function models; $^{**}$ indicates significant improvements at the $p < 0.001$ on paired significance t-test as compared to the baseline.} \label{tab:citation-function-evaluation-1}
\begin{tabular}{|c|c|c|c|c|}
\hline
\textbf{Model} & Baseline & CNN & MTL \\\hline \hline
\textbf{Precision} & 68.28\% & $68.78\% \pm 0.51\%$ & $69.55\% \pm 0.61\%$ \\\hline
\textbf{Recall} & 69.40\% & $68.65\% \pm 0.68\%$ & $72.33\% \pm 0.36\%$ \\\hline
\textbf{$F_1$} & 68.70\% & $68.31\% \pm 0.52\%$ & $69.63\% \pm 0.47\%^{**}$ \\\hline
\end{tabular}
\end{table*}

\begin{table*}
\centering
\caption{Performance of citation provenance models; $^{*}$ indicates significant improvements at the $p < 0.01$ on paired significance t-test as compared to the baseline.} \label{tab:citation-provenance-evaluation-1} 
\begin{tabular}{|c|c|c|c|}
\hline
\textbf{Model} & Baseline & dCNN & MTL \\\hline \hline
\textbf{Precision} & $71.82\%$ & $79.36\% \pm 1.71\%$ & $79.47\% \pm 1.37\%$ \\\hline
\textbf{Recall} & $72.13\%$ & $79.07\% \pm 1.84\%$ & $79.53\% \pm 1.36\%$ \\\hline
\textbf{$F_1$} & $71.68\%$ & $78.55\% \pm 1.67\%$ & $79.38\% \pm 1.36\%^{*}$ \\\hline
\end{tabular}
\end{table*}

\begin{table*}
\caption{Examples of citation function classifications.}
\label{tab:citfunc_evaluations_examples}
\begin{tabular}{|p{11cm}|p{1cm}|p{1.2cm}|p{0.8cm}|}
\hline
\textbf{Citing Sentence} & \textbf{Actual} & \textbf{Baseline} & \textbf{MTL} \\ \hline \hline
(1) This result is different from that in {(Wu and Wang, 2004)}, where their method achieved an error rate reduction of $21.96\%$ as compared with the method "Gen+Spec". & \textit{CoCo} & \textit{CoCo} & \textit{CoCo} \\ \hline
(2) We show that the performance of our approach (using simple lexical features) is comparable to that of the state-of-art statistical MT system {(Koehn et al., 2007)}. & \textit{CoCo} & \textit{Pos} & \textit{CoCo} \\ \hline
(3) Errors have been shown to have a significant impact on predicting learner level (Yannakoudakis et al., 2011). & \textit{Neut} & \textit{Weak} & \textit{Neut} \\ \hline
(4) {Georgescul et al. (2009)} note that while WindowDiff technically penalizes false positives and false negatives equally, false positives are in fact more likely; a false positive error occurs anywhere where there are more computed boundaries than boundaries in the reference, while a false negative error can only occur when a boundary is missed. & \textit{Neut} & \textit{Weak} & \textit{Neut} \\ \hline
(5) NETE mining from comparable corpora using phonetic mappings was proposed in {\bf (Tao et al., 2006)}, but the need for language specific knowledge restricts its applicability across languages. & {\it Weak} & {\it Neut} & {\it Neut} \\\hline
\end{tabular}
\end{table*}

\begin{table*}
\caption{Examples of citation provenance classifications.}
\label{tab:citprov_evaluations_examples}
\begin{tabular}{|p{5.3cm}|p{5.8cm}|p{1cm}|p{1.2cm}|p{1cm}|}
\hline
\textbf{Citing Sentence} & \textbf{Target Fragment} & \textbf{Actual} & \textbf{Baseline} & \textbf{dCNN} \\ \hline \hline
\small (1) Bigrams have recently been shown to be very successful features in supervised word sense disambiguation {(Pedersen, 2001)}. & \small This paper shows that the combination of a simple feature set made up of bigrams and a standard decision tree learning algorithm results in accurate word sense disambiguation. & \textit{Prov} & \textit{Prov} & \textit{Prov} \\ \hline
\small (2) However, detailed research {(Zhou et al., 2005)} shows that it is difficult to extract new effective features to further improve the extraction accuracy. & \small This suggests that feature-based methods can effectively combine different features from a variety of sources (\textit{e.g.} WordNet and gazetteers) that can be brought to bear on relation extraction. & \textit{Non-Prov} & \textit{Prov} & \textit{Non-Prov} \\ \hline
\small (3) A number of automatically acquired inference rule/paraphrase collections are available, such as {(Szpektor et al., 2004)}. & \small In this paper, we will propose an unsupervised method to discover paraphrases from a large untagged corpus. & \textit{Prov} & \textit{Non-Prov} & \textit{Prov} \\
\hline
\end{tabular}
\end{table*}

\subsection{Neural Models}
We implement both the supervised baselines in scikit-learn\footnote{Version 0.19.1.}; and the DNN models in Keras and Tensorflow\footnote{Keras 1.1.0, and Tensorflow 0.12.1.}. For deep learning training, GloVe word embedding are taken as feature representations of the tokens. For all the CNN-based neural models, we use typical settings: a window size of 5, a filter size of 256, a training minibatch size of 256, and a categorical cross entropy loss. We use averaged cross entropy loss over both tasks as the loss for the MTL model. For all neural models, we use RMSProp as the optimizer for 30 epochs.


As our data for citation function and citation provenance comes from different sources, we do not have an aligned training dataset such that for given citation contexts in the dataset, both function and provenance labels are available. This prevents us from training the complete MTL model in one go using both losses simultaneously. To mitigate this weakness in the training data, we use a batch-wise selective parameter training by calculating losses for each batch and tuning each halves of the network one at a time. 


\subsection{Evaluation}
We perform five-fold cross validation to evaluate all models. 
To appropriately separate train and test sets, 
during the cross-validation, 
provenance instances collected from the same article are grouped together, appearing all in either the training or test set. Since there is intrinsic randomness in Keras' implementation, we show neural model performance averaged over 5 runs. The standard deviation is indicated after the $\pm$ sign for each performance score. For both tasks, 
we report all models for precision, recall, and $F_1$ scores weighted over all classes. \\

\noindent{\bf Citation Function} \\
Table~\ref{tab:citation-function-evaluation-1} shows the performance of all models on the citation function dataset. 
We observe that the simple CNN model achieves at-par performance as the supervised baseline.   
MTL improves the supervised baseline by about 1\%, 
reaching around 70\%.  More specifically, 
we observe that these improvements could be attributed to both precision and recall. \\ 

\noindent{\bf Citation Provenance} \\
Table~\ref{tab:citation-provenance-evaluation-1} shows the precision, recall, and $F_1$ scores for our citation provenance experiments. We observe that the simple CNN model yields better results than the supervised baseline in all of the three evaluation measures. 
The MTL-learned dCNN model gives even better results with improvement over the plain CNN model by 1\%, bringing the scores up to about 79.4\% in all evaluation measures.   

\section{Discussion\label{sec:discussion}}
We now discuss and analyse the performance reached by the multi-task learning model, along with comparisons with the supervised baselines.

\subsection{Citation Function}
Table~\ref{tab:citfunc_evaluations_examples} shows some examples of function classifications. We observe that, 
in example (4), the citing sentence has multiple occurrences of words that have negative meanings  
such as ``false'', ``negative'', and ``error''. 
The neural model correctly captures the meaning of the whole sentence to produce a {\it Neut} classification and is not affected by these words. 
On the other hand, in example (5), the correct label is {\it Weak} but both the baseline and MTL 
incorrectly classify it as {\it Neut}. 
The reason why MTL fails to classify the citation sentence correctly is possibly due to the relative lack of explicit linguistic cues which indicate that the citing sentence is actually revealing a restriction of the cited work. 


\subsection{Citation Provenance}
Table~\ref{tab:citprov_evaluations_examples} shows some examples of provenance classifications obtained by the baseline and dCNN. 
Citing sentence (3) illustrates the advantages of a deep learning model. We observe that there are almost no word overlap between the two text inputs; this is likely why the baseline fails to classify it correctly, as it finds the two texts to be largely unrelated. 
 
However, closer examination of the sentences reveals the semantic similarity between them. For example, phrases such as ``automatically acquired'', ``inference'' share a common meaning as ``unsupervised''. Our deep learning model is able to classify the instance correctly as it has access to semantic relationships captured by the GloVe embeddings. 



We comment on some details of our implementation: 
for the neural models, we use only the citing sentence as input, 
instead of the citation context. This is because experiments with the citation context show degradation in performance: using the surrounding sentences may have accidentally introduced noise to the classification process. 
Furthermore, in choosing an appropriate neural model, we have also explored 
a number of alternatives, such as a Bidirectional Long Short-Term Memory (BiLSTM) network~\cite{Hochreiter97} for classification, double BiLSTM (similar to dCNN but with BiLSTM replacement) for provenance, and single CNN for provenance where both the citation context and candidate provenance text are appended and fed to a single CNN model. All the above mentioned models turn out to perform even worse that the feature-based baseline mostly cause of high number of parameters (in BiLSTM based models) and loss of fidelity (in case of merging citation context and provenance text).  
Our results show that CNN models are the most effective in our citation provenance prediction task.
Additionally, our models make use of local textual information only,
without global information such as \textit{Citing Location}, \textit{Year Difference}, etc.  Incorporating such information shows some gain.
However, for the sake of keeping the model domain-independent, we have eschewed the incorporation of any such features.

\section{Conclusion\label{sec:conclusion}}
We have investigated two  
related tasks in scholarly document processing, pertaining to citation analysis: citation function and provenance.  Our work outperforms 
existing state-of-the-arts for both tasks by applying 
a standard convolutional neural network architecture to the two tasks.  
We leverage our key insight of the relationship between the tasks and employ multi-task learning, 
resulting in further improvement in both tasks. 
We also contribute a citation function dataset and our code; these are released publicly to facilitate research replication and extension.

We conclude that better performance on these tasks will significantly enhance in-depth automated understanding on citations and their relation to scientific documents. Citations are not equally created as their functions, sentiments and scope are different. 
Continued work in this area will facilitate better measurement of the quality and impact of the scholarly literature. 


\bibliography{emnlp2018}

\begin{thebibliography}{22}
\expandafter\ifx\csname natexlab\endcsname\relax\def\natexlab#1{#1}\fi

\bibitem[{Abu-Jbara et~al.(2013)Abu-Jbara, Ezra, and Radev}]{abu2013purpose}
Amjad Abu-Jbara, Jefferson Ezra, and Dragomir Radev. 2013.
\newblock {Purpose and Polarity of Citation: Towards NLP-based Bibliometrics}.
\newblock In \emph{Proc. of the 2013 Conference of the North American Chapter
  of the Association for Computational Linguistics: Human Language Technologies
  (NAACL-HLT 2013)}, pages 596--606.

\bibitem[{Agarwal et~al.(2010)Agarwal, Choubey, and
  Yu}]{agarwal2010automatically}
Shashank Agarwal, Lisha Choubey, and Hong Yu. 2010.
\newblock {Automatically Classifying the Role of Citations in Biomedical
  Articles}.
\newblock In \emph{Proc. of American Medical Informatics Association Fall
  Symposium (AMIA 2010)}, pages 11--15.

\bibitem[{Alonso and Plank(2017)}]{Alonso17}
H.~M. Alonso and B.~Plank. 2017.
\newblock {When is multitask learning effective? Semantic sequence prediction
  under varying data conditions}.
\newblock In \emph{Proc. of the 15th Conference of the European Chapter of the
  Association for Computational Linguistics (EACL 2017)}, pages 44--53.

\bibitem[{Bingel and S{\o}gaard(2017)}]{Bingel17}
J.~Bingel and A.~S{\o}gaard. 2017.
\newblock {Identifying beneficial task relations for multi-task learning in
  deep neural networks}.
\newblock In \emph{Proc. of the 15th Conference of the European Chapter of the
  Association for Computational Linguistics (EACL 2017)}, pages 164--169.

\bibitem[{Bird et~al.(2008)Bird, Dale, Dorr, Gibson, Joseph, Kan, Lee, Powley,
  Radev, and Tan}]{bird2008acl}
Steven Bird, Robert Dale, Bonnie~J Dorr, Bryan~R Gibson, Mark~Thomas Joseph,
  Min-Yen Kan, Dongwon Lee, Brett Powley, Dragomir~R Radev, and Yee~Fan Tan.
  2008.
\newblock {The ACL Anthology Reference Corpus: A Reference Dataset for
  Bibliographic Research in Computational Linguistics}.
\newblock In \emph{Proc. of the 6th International Conference on Language
  Resources and Evaluation Conference (LREC'08)}, pages 1755--1759.

\bibitem[{Bromley et~al.(1994)Bromley, Guyon, LeCun, S{\"a}ckinger, and
  Shah}]{bromley1994signature}
Jane Bromley, Isabelle Guyon, Yann LeCun, Eduard S{\"a}ckinger, and Roopak
  Shah. 1994.
\newblock Signature verification using a" siamese" time delay neural network.
\newblock In \emph{Advances in neural information processing systems}, pages
  737--744.

\bibitem[{Caruana(1998)}]{caruana1998multitask}
Rich Caruana. 1998.
\newblock {Multitask Learning}.
\newblock \emph{Learning to Learn}, pages 95--133.

\bibitem[{Garzone and Mercer(2000)}]{garzone2000towards}
Mark Garzone and Robert~E Mercer. 2000.
\newblock {Towards an Automated Citation Classifier}.
\newblock In \emph{Proc. of 13th Biennial Conference of the Canadian Society
  for Computational Studies of Intelligence (AI 2000)}, Lecture Notes in
  Artificial Intelligence (LNAI), Vol. 1822, pages 337--346. Springer-Verlag.

\bibitem[{Hochreiter and Schmidhuber(1997)}]{Hochreiter97}
S.~Hochreiter and J.~Schmidhuber. 1997.
\newblock {Long Short-Term Memory}.
\newblock \emph{Neural Computation}, 9(8):1735--1780.

\bibitem[{Jaidka et~al.(2018)Jaidka, Chandrasekaran, Rustagi, and
  Kan}]{jaidka2018insights}
Kokil Jaidka, Muthu~Kumar Chandrasekaran, Sajal Rustagi, and Min-Yen Kan. 2018.
\newblock Insights from {CL-SciSumm} 2016: the faceted scientific document
  summarization shared task.
\newblock \emph{International Journal on Digital Libraries}, 19(2-3):163--171.

\bibitem[{Jha et~al.(2017)Jha, Jbara, Qazvinian, and Radev}]{jha2017nlp}
Rahul Jha, Amjad-Abu Jbara, Vahed Qazvinian, and Dragomir~R Radev. 2017.
\newblock {NLP}-driven citation analysis for scientometrics.
\newblock \emph{Natural Language Engineering}, 23(1):93--130.

\bibitem[{Low(2011)}]{wee2011citation}
Heng~Wee Low. 2011.
\newblock {Citation Provenance}.
\newblock \emph{Bachelor's thesis. School of Computing, National University of
  Singapore}.

\bibitem[{Moravcsik and Murugesan(1975)}]{moravcsik1975some}
Michael~J Moravcsik and Poovanalingam Murugesan. 1975.
\newblock {Some Results on the Function and Quality of Citations}.
\newblock \emph{Social Studies of Science}, 5(1):86--92.

\bibitem[{Munkhdalai et~al.(2016)Munkhdalai, Lalor, and
  Yu}]{munkhdalai2016citation}
Tsendsuren Munkhdalai, John Lalor, and Hong Yu. 2016.
\newblock Citation analysis with neural attention models.
\newblock In \emph{Proceedings of the Seventh International Workshop on Health
  Text Mining and Information Analysis}, pages 69--77.

\bibitem[{Plank et~al.(2016)Plank, S{\o}gaard, and Goldberg}]{Plank16}
B.~Plank, A.~S{\o}gaard, and Y.~Goldberg. 2016.
\newblock {Multilingual Part-of-Speech Tagging with Bidirectional Long
  Short-Term Memory Models and Auxiliary Loss}.
\newblock In \emph{Proc. of the 54th Annual Meeting of the Association for
  Computational Linguistics (ACL 2016)}, pages 412--418.

\bibitem[{Prasad(2017)}]{prasad:2017:CL-SciSumm}
Animesh Prasad. 2017.
\newblock {{WING-NUS} at {CL-SciSumm} 2017: Learning from Syntactic and
  Semantic Similarity for Citation Contextualization}.
\newblock In \emph{Proc. of the 2nd Joint Workshop on Bibliometric-enhanced
  Information Retrieval and Natural Language Processing for Digital Libraries
  (BIRNDL 2017)}, pages 26--32.

\bibitem[{Rei(2017)}]{rei:2017:Long}
Marek Rei. 2017.
\newblock {Semi-supervised Multitask Learning for Sequence Labeling}.
\newblock In \emph{Proc. of the 55th Annual Meeting of the Association for
  Computational Linguistics (ACL 2017)}, pages 2121--2130.

\bibitem[{Teufel et~al.(2006)Teufel, Siddharthan, and
  Tidhar}]{teufel2006automatic}
Simone Teufel, Advaith Siddharthan, and Dan Tidhar. 2006.
\newblock {Automatic Classification of Citation Function}.
\newblock In \emph{Proc. of the 2006 Conference on Empirical Methods in Natural
  Language Processing (EMNLP 2006)}, pages 103--110.

\bibitem[{{Times Higher Education}(2015)}]{citationaverages}
{Times Higher Education}. 2015.
\newblock Citation averages, 2000-2010, by fields and years.
\newblock \emph{Times Higher Education (THE)}.

\bibitem[{Wan et~al.(2009)Wan, Paris, Muthukrishna, and
  Dale}]{wan2009designing}
Stephen Wan, C{\'e}cile Paris, Michael Muthukrishna, and Robert Dale. 2009.
\newblock {Designing a citation-sensitive research tool: an initial study of
  browsing-specific information needs}.
\newblock In \emph{Proc. of the 2009 Workshop on Text and Citation Analysis for
  Scholarly Digital Libraries (NLPIR4DL '09)}, pages 45--53.

\bibitem[{Yin and Sch{\"u}tze(2015)}]{yin2015convolutional}
Wenpeng Yin and Hinrich Sch{\"u}tze. 2015.
\newblock Convolutional neural network for paraphrase identification.
\newblock In \emph{Proceedings of the 2015 Conference of the North American
  Chapter of the Association for Computational Linguistics: Human Language
  Technologies}, pages 901--911.

\bibitem[{Yulianto(2012)}]{yulianto2012citation}
Eric Yulianto. 2012.
\newblock {Citation Typing}.
\newblock \emph{Bachelor's thesis. School of Computing, National University of
  Singapore}.

\end{thebibliography}
\bibliographystyle{acl_natbib_nourl}

\end{document}